\documentclass[journal,comsoc]{IEEEtran}
\usepackage[T1]{fontenc}
\usepackage{cite}

\ifCLASSINFOpdf
   \usepackage[pdftex]{graphicx}
\else
   \usepackage[dvips]{graphicx}
\fi

\usepackage{amsmath}
\usepackage[cmintegrals]{newtxmath}

\usepackage[numbers]{natbib}
\usepackage{booktabs}
\usepackage{multirow}
\usepackage{multicol}
\usepackage{amsmath}
\usepackage{algorithm}
\usepackage{algorithmic}
\usepackage{mathtools}

\mathchardef\mhyphen="2D

\begin{document}
%
\title{Beyond Predefined Learning Objects: A Thinking-Learning Interaction Model for Up-to-Date Autonomous Robot Learning}

\author{Hong~Su
\IEEEcompsocitemizethanks{\IEEEcompsocthanksitem H. Su is with the School of Computer Science, Chengdu University of Information Technology, Chengdu, China.\\
 E-mail: suguest@126.com. \\
\protect\\
}
\thanks{}}

\markboth{Journal of \LaTeX\ Class Files,~Vol.~14, No.~8, August~2015}%
{Shell \MakeLowercase{\textit{et al.}}: Bare Demo of IEEEtran.cls for IEEE Communications Society Journals}
%

\maketitle

\begin{abstract}
Autonomous robots operating in open and changing environments cannot always rely on predefined inputs, outputs, and action routines. Although existing learning methods enable robots to improve their performance through environmental interaction, the objects of learning are often fixed in advance, such as input features, recognition outputs, network structures, task goals, or action sequences. This limits their ability to adapt when new features, new categories, or more efficient task routines appear during long-term operation. To address this problem, this paper proposes a thinking-learning interaction model for autonomous robots. The core idea is that thinking guides learning by identifying potential changes, selecting useful evidence, organizing training materials, and planning verification actions, while learning promotes thinking by updating task knowledge, feature-selection experience, action strategies, and future reasoning processes. Based on this bidirectional mechanism, the robot can gradually move beyond predefined learning settings and adapt its recognition relations and action relations through continuous interaction with the environment. Specifically, the proposed model supports adaptive input feature discovery, output category expansion, learning model update, and action routine reconstruction. Experimental results show that the proposed model improves the final recognition accuracy from 0.419 to 0.845 in feature adaptation, achieves higher new-category formation accuracy and model-update success rate, and reduces the average action length from 13.0 to 4.0 in action routine reconstruction. In learning-enhanced thinking, the useful evidence selection rate increases from 0.272 to 0.965, indicating that learning results can effectively improve future evidence selection and reasoning.
\end{abstract}

\begin{IEEEkeywords}
Autonomous robot learning, thinking-learning interaction, adaptive learning objects, Up-to-Date Learning.
\end{IEEEkeywords}

\IEEEpeerreviewmaketitle

 \section{Introduction}
\label{sec:introduction}

Autonomous robots \cite{liu2025llm} \cite{raptis2025agentic} are expected to operate in open and changing environments for a long period of time. In such environments, robots may encounter new objects, new task conditions, new environmental changes, and new operation routines that cannot be completely specified before deployment. Although existing learning-based methods allow robots to improve their performance through interaction with the environment, many of them still rely on predefined learning settings. For example, the input features, output categories, learning models, task goals, and action routines are often fixed in advance. The robot mainly learns model parameters or action policies within this predefined framework.

However, this assumption may not hold in open environments. A feature that is ignored at the beginning may later become important for recognition. A predefined output set may be insufficient when a new object type or task state appears. A learning model that is suitable for one type of task may become unsuitable when the learning object changes. A predefined action routine may also become inefficient when the robot repeatedly interacts with the same device or environment. For instance, when a robot operates a washing machine, it may initially rely on repeated observation and button pressing to find the quick-wash mode. After several successful interactions, it may discover that a shorter fixed action sequence can achieve the same goal. Similarly, in object recognition, a robot may initially distinguish objects using shape and weight, but later find that color or another feature provides a more effective clue in a specific environment.

These examples indicate that autonomous robot learning should not be limited to optimizing a fixed model under fixed inputs and outputs. Instead, the robot should be able to learn what should be learned and how the learning framework itself should be changed. In this paper, we refer to this ability as \emph{up-to-date learning}, where the robot continuously revises its learning objects according to newly observed environmental evidence and accumulated interaction experience. In other words, the learning objects themselves, including inputs, outputs, models, and action routines, should be adaptive and kept up to date.

To address this problem, this paper proposes a thinking-learning interaction model for up-to-date autonomous robot learning. The core idea is that thinking guides learning, and learning enhances thinking. On the one hand, the thinking process helps the robot identify possible limitations of its current learning objects, select useful evidence, organize learning materials, and plan verification actions. On the other hand, the learning results are used not only to improve task performance, but also to update the thinking process itself. Through this bidirectional interaction, the robot can gradually improve both its task knowledge and its ability to decide what should be learned next.

Different from a conventional learning pipeline, the proposed model treats learning as a closed-loop process. The robot first observes the environment and evaluates whether its current learning objects are sufficient. If the current inputs, outputs, models, or action routines are found to be incomplete or inefficient, the thinking module generates a learning plan. The robot then collects evidence from current observations, historical memory, or active interaction with the environment. Based on the collected evidence, the learning module produces candidate learning results, such as new features, new output categories, modified models, or reconstructed action routines. These results are verified before being accepted. Verified results are used to update the learning object set and are also stored as experience to improve future thinking.

The main contributions of this paper are summarized as follows.

\begin{itemize}
    \item A thinking-learning interaction model is proposed for autonomous robots in open environments. The model establishes a bidirectional mechanism in which thinking guides learning and learning improves thinking, enabling the robot to decide what should be learned and how future reasoning should be improved.

    \item The paper extends the object of robot learning from fixed model parameters to adaptive learning objects, including input features, output categories, learning models, action routines, and their relations. This allows the robot to move beyond predefined learning objects.

    \item A closed-loop update process is developed to support up-to-date learning in open environments. Through thinking-guided evidence collection, dynamic learning material construction, learning result verification, and thinking strategy improvement, the robot can continuously revise its learning objects and keep its recognition and action relations up to date during long-term interaction.
\end{itemize}

The remainder of this paper is organized as follows. Section~\ref{sec:related_work} reviews related work. Section~\ref{sec:model} presents the proposed thinking-learning interaction model. Section~\ref{sec:up_to_date_adaptation} describes up-to-date adaptation of learning objects. Section~\ref{sec:verification} reports the verification results. Section~\ref{sec:conclusion} concludes the paper.

\section{Related Work}
\label{sec:related_work}

\subsection{Continual Learning and Open-World Recognition}
\label{subsec:continual_open_world}

Continual learning aims to enable learning systems to acquire new knowledge from non-stationary data streams without forgetting previously learned knowledge. Existing studies have investigated important problems such as catastrophic forgetting, memory replay, transfer learning, structural plasticity, and the stability-plasticity trade-off~\cite{parisi2019continual,delange2021continual}. These studies provide an important foundation for long-term learning systems, especially when new data arrive sequentially.

Open-world recognition further considers the problem that new and unknown categories may appear after deployment. Bendale and Boult~\cite{bendale2015open} formally defined open-world recognition and emphasized that a recognition system should detect unknown classes and incrementally add new categories. This direction is closely related to the output adaptation problem in this paper.

However, most continual learning and open-world recognition methods still assume that the learning framework is largely predefined. For example, the input representation, model type, learning procedure, and adaptation logic are usually determined before learning begins. In contrast, this paper focuses not only on learning new knowledge within a predefined setting, but also on adapting the learning objects themselves.

\subsection{Robot Learning and Autonomous Data Collection}
\label{subsec:robot_learning_data_collection}

Robot learning studies how robots acquire perception, control, and decision-making abilities through interaction with the physical environment. Reinforcement learning and imitation learning have been widely used to improve robotic policies.

However, real-world robot learning often requires costly data collection, carefully designed environments, reset mechanisms, success detectors, or human demonstrations. Recent work on autonomous robot data collection has attempted to reduce human supervision by allowing robots to collect data more independently~\cite{mirchandani2024scale}. AutoRT also shows that foundation models can help orchestrate multiple robots to collect diverse real-world data in unseen environments~\cite{ahn2024autort}.

These studies show that autonomous data collection is important for scalable robot learning. Nevertheless, the main focus is often on collecting more useful data for training existing policies or models. The proposed model emphasizes a different but complementary problem: how the robot decides what evidence should be collected and how the collected evidence should change the learning framework itself.

\subsection{Large Language Models for Robotic Reasoning and Planning}
\label{subsec:llm_robotics}

Large language models \cite{naveed2025comprehensive} have recently been applied to robotic task planning, instruction understanding, decision making, and embodied control. SayCan combines language model knowledge with affordance-based value functions so that high-level language plans can be grounded in feasible robotic actions~\cite{ahn2022saycan}. Code as Policies uses language models to generate robot policy code from natural language commands, enabling robots to compose control logic and use external libraries for spatial and geometric reasoning~\cite{liang2023code}. Recent surveys also show that LLMs are increasingly used in robot perception, planning, control, and human-robot interaction~\cite{kim2024llmrobotics,zeng2023llmrobotics}.

These works demonstrate that LLMs can provide strong reasoning and planning abilities for robots. However, many LLM-based robotic systems mainly use the language model as a planner, controller, or code generator. The thinking module in this paper has a different role. It does not only generate plans for current tasks, but also guides learning by identifying what should be learned, selecting evidence, constructing learning materials, and verifying candidate updates.

\subsection{Vision-Language-Action Models and Open-Ended Embodied Agents}
\label{subsec:vla_open_ended_agents}

Vision-language-action models integrate visual perception, language understanding, and action generation into robotic control. RT-2 shows that web-scale vision-language knowledge can be transferred to robotic control by expressing actions as tokens and co-fine-tuning vision-language models with robotic trajectory data~\cite{brohan2023rt2}. Such models improve generalization and enable robots to respond to novel objects and language instructions.

Open-ended embodied agents also provide useful inspiration for long-term autonomous learning. Voyager, for example, uses a large language model to build an embodied lifelong learning agent in Minecraft. It uses an automatic curriculum, a skill library, and iterative prompting with environmental feedback to acquire and reuse skills~\cite{wang2023voyager}. These methods show the value of skill accumulation, memory, and feedback-driven improvement.

However, these studies mainly focus on improving embodied task performance, skill acquisition, or generalization. The proposed model focuses on a more explicit thinking-learning interaction mechanism. It treats the learning model, input features, output categories, and action routines as adaptive objects. It also treats the thinking process itself as something that can be improved by learning.

\subsection{Comparison with Existing Studies}
\label{subsec:related_comparison}

The above studies provide important foundations for autonomous robot learning in open environments. Continual learning and open-world recognition address incremental knowledge acquisition and unknown category discovery. Robot learning and autonomous data collection address how robots can obtain experience from physical interaction. LLM-based robotics provides reasoning and planning abilities, while VLA models and open-ended agents show promising directions for embodied generalization and skill accumulation.

However, these directions usually emphasize one part of the problem: learning new classes, collecting data, generating plans, or acquiring skills. In contrast, this paper focuses on the interaction between thinking and learning. The proposed model argues that a robot should not only learn within predefined inputs, outputs, models, and action routines, but also learn how these learning objects should be changed. Thinking guides evidence collection and learning material construction, while learning results update both task knowledge and future thinking strategies.

\section{Thinking-Learning Interaction Model}
\label{sec:model}

\subsection{Overview}
\label{subsec:model_overview}

Autonomous robots operating in open environments are often required to handle situations that cannot be fully specified before deployment. In many existing learning-based systems, the input features, output categories, task goals, learning models, and action routines are defined in advance. The robot then learns within this predefined space. However, when the environment changes over time, such predefined settings may become insufficient. New features may become useful for recognition, new outputs may need to be added, models may need to be updated, and more efficient action routines may be discovered through long-term interaction.

To address this problem, this paper proposes a thinking-learning interaction model. The basic idea is that learning should not be treated as an isolated process driven only by fixed training data or predefined rewards. Instead, learning is guided by a thinking process that identifies what should be learned, why it should be learned, and how useful evidence should be collected. At the same time, the results of learning are used to improve the thinking process itself, enabling the robot to reason more effectively in future tasks.

The proposed model contains two closely connected directions. The first direction is \emph{thinking-guided learning}. In this process, the thinking module observes the current task state, compares it with previous experience, discovers possible limitations of existing inputs, outputs, models, or action routines, and then guides the collection of new evidence. Based on the collected evidence, the robot can learn new features, expand output categories, update learning models, or reconstruct action sequences.

The second direction is \emph{learning-enhanced thinking}. After the robot obtains new learning results, these results are not only used for the current task but are also stored as experience for future reasoning. For example, if a newly discovered feature repeatedly helps distinguish objects, the thinking module can give higher priority to similar features in later situations. If a newly learned action sequence completes a task more efficiently, the robot can use this experience to guide future action planning.

Therefore, the proposed model forms a closed-loop process:
\begin{equation}
\begin{aligned}
\text{Thinking}
&\rightarrow \text{Evidence Collection} \\
&\rightarrow \text{Learning}
\rightarrow \text{Thinking Update}.
\end{aligned}
\end{equation}

Through this loop, the robot gradually moves beyond predefined learning objects. The model does not assume that all useful features, categories, models, or task procedures are known in advance. Instead, it allows the robot to continuously adjust its learning objects according to environmental feedback and accumulated experience.

\subsection{Problem Definition}
\label{subsec:problem_definition}

Let an autonomous robot operate in an open environment over a sequence of time steps $t=1,2,\ldots,T$. At each time step, the robot receives observations from the environment, executes actions, obtains feedback, and updates its internal knowledge. Unlike a closed or fully predefined environment, the open environment considered in this paper may contain changing objects, changing task conditions, newly appearing features, unknown output categories, and alternative action routines.

In conventional learning-based robot systems, the learning process is usually defined within a fixed setting:
\begin{equation}
Y = f_{\theta}(X),
\end{equation}
where $X$ denotes the predefined input feature space, $Y$ denotes the predefined output space, and $f_{\theta}$ denotes a learning model with parameters $\theta$. The learning process mainly focuses on optimizing $\theta$ based on available training data.

However, in open environments, the limitation is not only whether the model parameters are well trained. The predefined learning objects themselves may become insufficient. Therefore, the robot needs to adapt not only the model parameters but also the learning objects. In this paper, the learning object set is defined as
\begin{equation}
\mathcal{O}_t = \{X_t, Y_t, M_t, A_t, R_t\},
\end{equation}
where $X_t$ is the input feature set at time $t$, $Y_t$ is the output or task-result set, $M_t$ is the current learning model, $A_t$ is the available action routine set, and $R_t$ represents the learned relation among inputs, outputs, models, and actions.

The adaptation problem can therefore be described as
\begin{equation}
\mathcal{O}_{t+1}
=
\operatorname{Adapt}(\mathcal{O}_t, E_t, K_t, \Phi_t),
\end{equation}
where $E_t$ denotes newly collected evidence from the environment, $K_t$ denotes the robot's accumulated knowledge and memory, and $\Phi_t$ denotes the current thinking strategy. The problem studied in this paper is how the robot uses thinking to guide evidence collection and learning, and then uses learning results to improve future thinking, so that it can continuously adapt its learning objects in an open environment.

The objective is to obtain a sequence of updated learning objects and thinking strategies:
\begin{equation}
(\mathcal{O}_0, \Phi_0)
\rightarrow
(\mathcal{O}_1, \Phi_1)
\rightarrow
\cdots
\rightarrow
(\mathcal{O}_T, \Phi_T).
\end{equation}

\subsection{Thinking-Learning Interaction Framework}
\label{subsec:thinking_learning_framework}

Based on the above problem definition, this paper constructs a thinking-learning interaction framework to support adaptive robot learning in open environments. At time step $t$, the robot maintains a thinking strategy $\Phi_t$, a knowledge state $K_t$, and a set of learning objects $\mathcal{O}_t$. When the robot interacts with the environment, it obtains an observation $s_t$ and evaluates whether the current learning objects are sufficient for the task.

The thinking process can be represented as
\begin{equation}
P_t = \operatorname{Think}(s_t, \mathcal{O}_t, K_t, \Phi_t),
\end{equation}
where $P_t$ denotes the learning plan generated by the thinking module. The plan may include the candidate feature to be checked, the possible new output category to be verified, the model modification to be considered, the action routine to be tested, and the type of evidence that should be collected.

According to $P_t$, the robot collects evidence from the current environment, historical memory, or additional interactions:
\begin{equation}
E_t = \operatorname{Collect}(P_t, s_t, K_t, \mathcal{A}_t),
\end{equation}
where $\mathcal{A}_t$ denotes the available actions for exploration and verification. The collected evidence is then used by the learning module:
\begin{equation}
L_t = \operatorname{Learn}(E_t, \mathcal{O}_t, K_t),
\end{equation}
where $L_t$ represents the newly obtained learning result. It may be a new feature relation, a new output category, a revised model, or a more efficient action sequence.

After learning, the robot updates its learning objects:
\begin{equation}
\mathcal{O}_{t+1}
=
\operatorname{UpdateObject}(\mathcal{O}_t, L_t).
\end{equation}
At the same time, the learning result also updates the thinking strategy:
\begin{equation}
\Phi_{t+1}
=
\operatorname{UpdateThinking}(\Phi_t, L_t, K_t).
\end{equation}

The overall interaction framework can therefore be summarized as
\begin{equation}
(s_t, \mathcal{O}_t, K_t, \Phi_t)
\rightarrow
P_t
\rightarrow
E_t
\rightarrow
L_t
\rightarrow
(\mathcal{O}_{t+1}, \Phi_{t+1}, K_{t+1}).
\end{equation}

\subsubsection{Thinking-Guided Learning}
\label{subsec:thinking_guided_learning}

Thinking-guided learning refers to the process in which the robot uses a thinking module to determine what should be learned, which evidence should be collected, and how the learning process should be organized. In open environments, the robot may encounter situations that cannot be well handled by its current learning objects. These situations may be caused by missing input features, insufficient output categories, unsuitable models, ineffective recognition relations, or inefficient action routines.

At time step $t$, the robot first evaluates whether the current learning objects $\mathcal{O}_t$ can explain the observed situation or support the current task:
\begin{equation}
q_t = \operatorname{Evaluate}(s_t, \mathcal{O}_t, K_t).
\end{equation}
If $q_t$ indicates that the current learning objects are sufficient, the robot can continue using the existing model or routine. Otherwise, the thinking module is activated:
\begin{equation}
I_t = \operatorname{IdentifyNeed}(q_t, s_t, \mathcal{O}_t, K_t),
\end{equation}
where $I_t$ denotes the identified learning need.

After the learning need is identified, the thinking module generates an evidence collection plan:
\begin{equation}
P_t = \operatorname{PlanEvidence}(I_t, \mathcal{O}_t, K_t, \Phi_t).
\end{equation}
The evidence may come from current observations, historical memory, additional sensing, or active interaction with the environment. The collected evidence is then used to construct learning materials:
\begin{equation}
D_t = \operatorname{ConstructData}(E_t, I_t, K_t),
\end{equation}
where $D_t$ denotes the learning materials constructed from evidence $E_t$.

Based on $D_t$, the learning module produces a learning result:
\begin{equation}
L_t = \operatorname{Learn}(D_t, \mathcal{O}_t).
\end{equation}
The thinking module further checks whether the learning result is reliable:
\begin{equation}
v_t = \operatorname{Verify}(L_t, E_t, K_t),
\end{equation}
where $v_t$ denotes the verification result. Therefore, thinking-guided learning can be summarized as
\begin{equation}
s_t \rightarrow q_t \rightarrow I_t \rightarrow P_t
\rightarrow E_t \rightarrow D_t \rightarrow L_t \rightarrow v_t.
\end{equation}

\subsubsection{Learning-Enhanced Thinking}
\label{subsec:learning_enhanced_thinking}

Learning-enhanced thinking refers to the process in which the robot uses learning results to improve its future thinking strategy. In the proposed model, learning is not only used to improve task performance, but also used to update how the robot identifies learning needs, selects evidence, organizes learning materials, evaluates candidate features, and plans future actions.

After the robot obtains a learning result $L_t$, the result is evaluated according to its usefulness for the current task and its reliability in the environment:
\begin{equation}
u_t = \operatorname{Assess}(L_t, v_t, s_t, K_t),
\end{equation}
where $v_t$ is the verification result, $s_t$ is the current observation, $K_t$ is the accumulated knowledge, and $u_t$ denotes the usefulness of the learning result. The evaluated learning result is then stored into the knowledge state:
\begin{equation}
K_{t+1} = \operatorname{UpdateKnowledge}(K_t, L_t, u_t).
\end{equation}

The updated knowledge state includes not only task-related knowledge, such as new features, new output categories, or new action routines, but also meta-level experience about the thinking process. Based on the updated knowledge, the thinking strategy is revised as
\begin{equation}
\Phi_{t+1}
=
\operatorname{ImproveThinking}(\Phi_t, K_{t+1}, u_t).
\end{equation}

More importantly, the thinking process itself can be regarded as a learnable object. In the early stage, the robot may have little experience about how to select useful learning materials. In this case, it may randomly select candidate features, evidence sources, or historical fragments, and then verify whether they are useful for the current learning task. Through repeated selection, verification, success, and failure, the robot gradually accumulates experience about how to select better learning materials in later situations.

This process is similar to human experience-based material collection. When direct evidence is insufficient, the robot may learn to expand the evidence collection scope. For example, it may search related historical observations, recall previous similar tasks, or retrieve partial records that share similar inputs, outputs, task goals, action patterns, or environmental conditions. Such scope expansion is not only a predefined rule, but can also become a learned thinking strategy.

The improvement of thinking can therefore be expressed as a feedback process:
\begin{equation}
L_t \rightarrow u_t \rightarrow K_{t+1} \rightarrow \Phi_{t+1}.
\end{equation}

If a large language model is used as the thinking module, the accumulated thinking experience can be used for retraining or fine-tuning the model. Alternatively, the experience can be stored as memory and retrieved as contextual material during later thinking.

\subsection{Closed-Loop Update Process}
\label{subsec:closed_loop_update}

The proposed thinking-learning interaction model works as a closed-loop update process. At time step $t$, the robot receives an observation $s_t$ from the environment and uses the current learning object set $\mathcal{O}_t$, knowledge state $K_t$, and thinking strategy $\Phi_t$ to evaluate the current situation:
\begin{equation}
q_t = \operatorname{Evaluate}(s_t, \mathcal{O}_t, K_t).
\end{equation}

If the current learning objects are sufficient, the robot continues to use the existing recognition relation, model, or action routine. If the evaluation result indicates that the current learning objects are incomplete, unreliable, or inefficient, the thinking module generates a learning plan:
\begin{equation}
P_t = \operatorname{Think}(q_t, s_t, \mathcal{O}_t, K_t, \Phi_t).
\end{equation}

The robot collects evidence:
\begin{equation}
E_t = \operatorname{Collect}(P_t, s_t, K_t, \mathcal{A}_t),
\end{equation}
constructs learning materials:
\begin{equation}
D_t = \operatorname{ConstructData}(E_t, P_t, K_t),
\end{equation}
and learns from these materials:
\begin{equation}
L_t = \operatorname{Learn}(D_t, \mathcal{O}_t).
\end{equation}

The learning result is verified before being accepted:
\begin{equation}
v_t = \operatorname{Verify}(L_t, E_t, K_t).
\end{equation}
If $v_t$ is positive, the learning object set is updated:
\begin{equation}
\mathcal{O}_{t+1}
=
\operatorname{UpdateObject}(\mathcal{O}_t, L_t, v_t).
\end{equation}
The robot also updates its knowledge state and thinking strategy:
\begin{equation}
K_{t+1}
=
\operatorname{UpdateKnowledge}(K_t, L_t, v_t),
\end{equation}
\begin{equation}
\Phi_{t+1}
=
\operatorname{ImproveThinking}(\Phi_t, K_{t+1}, L_t, v_t).
\end{equation}

The complete closed-loop process can be summarized as
\begin{equation}
\begin{aligned}
&(s_t, \mathcal{O}_t, K_t, \Phi_t)
\rightarrow q_t
\rightarrow P_t
\rightarrow E_t
\rightarrow D_t \\
&\rightarrow L_t
\rightarrow v_t
\rightarrow
(\mathcal{O}_{t+1}, K_{t+1}, \Phi_{t+1}).
\end{aligned}
\end{equation}

\subsubsection{Algorithm}
\label{subsec:algorithm}

Based on the above closed-loop update process, the proposed thinking-learning interaction model can be described as Algorithm~\ref{alg:thinking_learning}.

\begin{algorithm}[htpb]
\caption{Thinking-Learning Interaction Model}
\label{alg:thinking_learning}
\begin{algorithmic}[1]
\REQUIRE Initial learning object set $\mathcal{O}_0$, initial knowledge state $K_0$, initial thinking strategy $\Phi_0$, environment $\mathcal{E}$
\ENSURE Updated learning objects $\mathcal{O}_T$, knowledge state $K_T$, and thinking strategy $\Phi_T$

\FOR{$t=0$ to $T-1$}
    \STATE Obtain current observation $s_t$ from the environment $\mathcal{E}$.
    \STATE Evaluate whether the current learning objects are sufficient:
    \[
    q_t = \operatorname{Evaluate}(s_t, \mathcal{O}_t, K_t).
    \]
    \IF{$q_t$ indicates that no adaptation is required}
        \STATE Execute the current recognition relation, model, or action routine.
        \STATE Set $\mathcal{O}_{t+1} = \mathcal{O}_t$.
        \STATE Update $K_{t+1}$ with the new observation.
        \STATE Set $\Phi_{t+1} = \Phi_t$.
    \ELSE
        \STATE Generate a learning plan:
        \[
        P_t = \operatorname{Think}(q_t, s_t, \mathcal{O}_t, K_t, \Phi_t).
        \]
        \STATE Collect evidence:
        \[
        E_t = \operatorname{Collect}(P_t, s_t, K_t, \mathcal{A}_t).
        \]
        \STATE Construct learning materials:
        \[
        D_t = \operatorname{ConstructData}(E_t, P_t, K_t).
        \]
        \STATE Learn from the constructed materials:
        \[
        L_t = \operatorname{Learn}(D_t, \mathcal{O}_t).
        \]
        \STATE Verify the learning result:
        \[
        v_t = \operatorname{Verify}(L_t, E_t, K_t).
        \]
        \IF{$v_t$ is positive}
            \STATE Update the learning object set:
            \[
            \mathcal{O}_{t+1}
            =
            \operatorname{UpdateObject}(\mathcal{O}_t, L_t, v_t).
            \]
        \ELSE
            \STATE Reject the candidate result or mark it for further evidence collection.
            \STATE Set $\mathcal{O}_{t+1} = \mathcal{O}_t$.
        \ENDIF
        \STATE Update the knowledge state:
        \[
        K_{t+1}
        =
        \operatorname{UpdateKnowledge}(K_t, L_t, v_t).
        \]
        \STATE Improve the thinking strategy:
        \[
        \Phi_{t+1}
        =
        \operatorname{ImproveThinking}(\Phi_t, K_{t+1}, L_t, v_t).
        \]
    \ENDIF
\ENDFOR

\RETURN $\mathcal{O}_T$, $K_T$, $\Phi_T$
\end{algorithmic}
\end{algorithm}

\section{Up-to-Date Adaptation of Learning Objects}
\label{sec:up_to_date_adaptation}

\subsection{Overview of Up-to-Date Learning in Open Environments}
\label{subsec:up_to_date_learning_overview}

In this paper, \emph{up-to-date learning} refers to the ability of an autonomous robot to continuously revise its learning objects according to newly observed environmental evidence and accumulated interaction experience. Different from conventional learning that mainly updates model parameters within a predefined framework, up-to-date learning updates the framework itself, including input features, output categories, learning models, action routines, and their relations.

In open environments, autonomous robots should not regard their learning objects as permanently fixed. When the environment changes, the robot may need to update not only model parameters, but also the input features, output categories, model structures, action routines, and relations among them. For example, a robot may discover that a previously ignored feature becomes useful for recognition, that an unknown object cannot be represented by existing output categories, that the current model is unsuitable for a changed learning target, or that a shorter action routine can complete the same task more efficiently.

The proposed thinking-learning interaction model provides the basic mechanism for this process. Thinking identifies whether the current learning objects are insufficient or outdated. Learning uses collected evidence to generate candidate updates. Verification determines whether these updates should be accepted. Once accepted, the updated learning objects become part of the robot's current learning framework and influence future reasoning.

Thus, up-to-date learning means that the robot continuously aligns its learning objects with the changing environment. It does not wait for a human designer to manually redefine inputs, outputs, models, or routines. Instead, it uses long-term interaction to discover what should be updated and how the update should be validated.

\subsection{Mechanism of Up-to-Date Learning}
\label{subsec:up_to_date_learning_mechanism}

Up-to-date learning is triggered when the robot finds that its current learning objects cannot sufficiently explain new observations or support efficient task execution. This may occur when the recognition result becomes unstable, when repeated errors appear under the current feature set, when unknown samples cannot be represented by existing output categories, when the current model is unsuitable for the changed learning target, or when a predefined action routine becomes inefficient. In addition, up-to-date learning may also be triggered when the robot observes new indicators, features, or clues that are more salient, more stable, or more adaptive than the currently used learning objects. In this case, the current learning objects may still be usable, but they are no longer the most suitable choice for the changing environment.

Let the current learning object set be

\begin{equation}
\mathcal{O}_t = \{X_t, Y_t, M_t, A_t, R_t\},
\end{equation}

where $X_t$ denotes the input feature set, $Y_t$ denotes the output or task-result set, $M_t$ denotes the learning model, $A_t$ denotes the action routine set, and $R_t$ denotes the relations among inputs, outputs, models, and actions. The purpose of up-to-date learning is to determine whether any component in $\mathcal{O}_t$ should be revised according to newly observed evidence and accumulated experience.

The possible component requiring update can be detected as

\begin{equation}
C_t = \operatorname{DetectUpdateNeed}(\mathcal{O}_t, s_t, K_t),
\end{equation}

where $s_t$ is the current observation, $K_t$ is the accumulated knowledge state, and $C_t$ denotes the component that may require revision. For example, $C_t$ may correspond to the input feature set if the current features cannot distinguish similar objects or if newly observed features provide clearer and more adaptive evidence. It may correspond to the output set if an unknown category repeatedly appears, to the model if the current network is no longer suitable for the updated learning object, or to the action routine if the current routine is inefficient or a more direct routine is discovered.

After a component requiring update is detected, the robot does not revise it immediately. Instead, the thinking module generates an evidence collection and verification plan:

\begin{equation}
P_t = \operatorname{PlanUpdate}(C_t, \mathcal{O}_t, K_t, \Phi_t),
\end{equation}

where $\Phi_t$ is the current thinking strategy. The plan specifies what evidence should be collected, where the evidence may come from, and how the candidate update should be verified. For example, if a new feature is found to be more salient than the current features, the plan may require the robot to collect additional samples under different conditions to check whether this feature is consistently useful. If a shorter action routine is discovered, the plan may require repeated execution to verify whether it can reliably achieve the same task goal.

The evidence can come from current observations, historical memory, additional sensing, or active interaction with the environment:

\begin{equation}
E_t = \operatorname{Collect}(P_t, s_t, K_t, \mathcal{A}_t),
\end{equation}

where $\mathcal{A}_t$ denotes the available actions for exploration and verification. The collected evidence is then used to generate a candidate update:

\begin{equation}
\Delta \mathcal{O}_t = \operatorname{LearnUpdate}(E_t, C_t, \mathcal{O}_t).
\end{equation}

The candidate update may add a new input feature, replace a weak feature, expand the output set, modify the learning model, reconstruct an action routine, or update the relation among them. For example, if color is repeatedly verified as a clearer feature for distinguishing two objects than shape, the candidate update may add color into $X_t$ or increase its priority in the recognition relation. If an unknown object repeatedly appears and cannot be explained by existing categories, the candidate update may add a new output category to $Y_t$. If a shorter action sequence reaches the same goal with higher stability, the candidate update may revise $A_t$.

Since open environments may contain noise, temporary changes, and accidental correlations, the candidate update should not be accepted only because it appears once. Therefore, the robot verifies the candidate update before using it:

\begin{equation}
v_t = \operatorname{Verify}(\Delta \mathcal{O}_t, E_t, K_t).
\end{equation}

If $v_t$ is positive, the learning object set is updated as

\begin{equation}
\mathcal{O}_{t+1}
=
\mathcal{O}_t \oplus \Delta \mathcal{O}_t,
\end{equation}

where $\oplus$ denotes a revision operation on the learning object set. If $v_t$ is negative, the candidate update is rejected or stored as a candidate for future verification. This design prevents the robot from treating a single abnormal observation, temporary variation, or accidental correlation as stable knowledge.

This mechanism distinguishes up-to-date learning from simple online parameter updating. Online updating mainly changes model parameters under a fixed learning framework, whereas up-to-date learning revises the learning framework itself. It keeps the robot's learning objects aligned with the current environment by continuously detecting update needs, collecting evidence, verifying candidate updates, and revising the learning object set.

Therefore, up-to-date learning is both adaptive and conservative. It is adaptive because the robot can update its learning objects when new features, categories, models, or action routines become more suitable for the environment. It is conservative because an update is accepted only after evidence collection and verification. This balance allows the robot to remain responsive to open environments while avoiding unnecessary or unreliable changes.

\subsection{Adaptation of Learning Objects}
\label{subsec:adaptation_objects}

Based on the above mechanism, the robot can adapt four important types of learning objects: input features, output categories, learning models, and action routines. These objects are often predefined in conventional robot learning systems. However, in open environments, they may become incomplete, unsuitable, or inefficient.

First, the robot can adapt its input features. Let the input feature set at time $t$ be denoted as

\begin{equation}
X_t = \{x_1, x_2, \ldots, x_n\}.
\end{equation}

In a predefined learning system, the model usually learns a mapping from $X_t$ to the output set $Y_t$. However, some useful features may not be included in the initial input set. When the thinking module finds that the current features cannot sufficiently distinguish objects or task states, it can generate candidate features from current observations or historical memory. If a candidate feature $x_{\mathrm{new}}$ is verified as useful, the input feature set is updated as

\begin{equation}
X_{t+1} = X_t \cup \{x_{\mathrm{new}}\}.
\end{equation}

For example, if a robot originally distinguishes two objects based on shape and weight, but later finds that color is a more effective feature in a specific environment, color can be added as a new input feature. In this case, the robot does not simply optimize the parameters of the original model, but changes the input space of the learning problem.

Second, the robot can adapt its output categories or task-result set. Let the output set at time $t$ be

\begin{equation}
Y_t = \{y_1, y_2, \ldots, y_m\}.
\end{equation}

In open environments, the robot may encounter observations that cannot be properly explained by existing outputs. If the current output set cannot represent a newly encountered object, state, or task result, the thinking module can mark it as a candidate new category. After sufficient evidence is collected and verified, the output set can be expanded as

\begin{equation}
Y_{t+1} = Y_t \cup \{y_{\mathrm{new}}\}.
\end{equation}

This process allows the robot to move beyond fixed output definitions. A new category is not added only because a single abnormal observation appears. Instead, it is added after repeated evidence collection, comparison with existing categories, and verification by the thinking module or external feedback. This reduces the risk of treating noise or temporary variation as a new output category.

Third, the robot can adapt its learning model. Let the learning model at time $t$ be denoted as $M_t$. In conventional systems, the model type and structure are usually selected before deployment. For example, a CNN may be used for image recognition, an LSTM may be used for sequence prediction, and a Transformer may be used for long sequence modeling. However, when the learning object changes, the original model may no longer be suitable. For example, if the task changes from static image recognition to action sequence prediction, a sequence model such as an LSTM or Transformer may be more appropriate than a CNN. If new visual features or new output categories are discovered, the robot may need to modify the input layer, add a new output head, fine-tune the current network, or replace the model with another architecture.

The model adaptation process can be represented as

\begin{equation}
M_{t+1}
=
\operatorname{UpdateModel}(M_t, L_t, v_t, \mathcal{O}_t),
\end{equation}

where $L_t$ denotes the learning result and $v_t$ denotes the verification result. The update of $M_t$ may include retraining the current model, fine-tuning a pretrained model, modifying the input layer, expanding the output layer, selecting another model type, or replacing the current model with a more suitable architecture.

Fourth, the robot can adapt its action routines. Let an action routine at time $t$ be represented as

\begin{equation}
A_t = (a_1, a_2, \ldots, a_k),
\end{equation}

where each $a_i$ is an executable action. In many robotic systems, a task is completed by following a predefined sequence of actions. However, such a routine may be inefficient, redundant, or invalid when the environment changes. Through long-term interaction, the robot may discover that another sequence can achieve the same goal with fewer actions or higher reliability. Therefore, the action routine can be updated as

\begin{equation}
A_{t+1} = (a'_1, a'_2, \ldots, a'_l),
\end{equation}

where the new sequence may be shorter, more stable, or more suitable for the current environment.

For example, when operating a washing machine, a robot may initially use a repeated observation-and-press routine to find the quick-wash mode. After several successful trials, it may learn that pressing the program button a fixed number of times after power-on can directly reach the target mode. In this case, the robot reconstructs the action routine from a reactive process into a more efficient learned sequence.

The adaptation of these four objects can be described uniformly by updating the learning object set:

\begin{equation}
\mathcal{O}_{t+1}
=
\operatorname{UpdateObject}(\mathcal{O}_t, L_t, v_t).
\end{equation}

When $X_t$, $Y_t$, $M_t$, or $A_t$ changes, the corresponding relations in $R_t$ should also be updated. Therefore, the adaptation of learning objects is not an isolated update of individual components, but a reconstruction of the robot's current learning framework.

\subsection{Formation of New Recognition and Action Relations}
\label{subsec:new_relations}

Up-to-date adaptation does not only change isolated learning objects. More importantly, it updates the relations among them. In the proposed model, two types of relations are especially important: the recognition relation and the action relation.

The recognition relation can be represented as

\begin{equation}
R_t^{rec}: (X_t, M_t) \rightarrow Y_t,
\end{equation}

where $X_t$ is the current input feature set, $M_t$ is the current learning model, and $Y_t$ is the current output set. This relation describes how the robot maps observed features to recognition results or task states. When a new feature is added, a new category is introduced, or the model is changed, the recognition relation should be reconstructed.

The update of the recognition relation can be represented as

\begin{equation}
R_{t+1}^{rec}
=
\operatorname{UpdateRelation}(R_t^{rec}, X_{t+1}, M_{t+1}, Y_{t+1}).
\end{equation}

This means that the robot should not simply add a new feature or category independently. Instead, it should verify whether the updated feature set, model, and output set form a more reliable recognition relation.

The action relation can be represented as

\begin{equation}
R_t^{act}: A_t \rightarrow G_t,
\end{equation}

where $A_t$ denotes the current action routine and $G_t$ denotes the task goal or task result. This relation describes how an action sequence leads to task completion. When the robot discovers a shorter or more reliable action sequence, the action relation is updated as

\begin{equation}
R_{t+1}^{act}
=
\operatorname{UpdateActionRelation}(R_t^{act}, A_{t+1}, G_t).
\end{equation}

For example, in a device operation task, the original relation may be based on repeated observation and button pressing. After long-term interaction, the robot may discover a direct action routine that achieves the same goal. In this case, the new action relation is more efficient while still being verified by task success.

Therefore, the proposed adaptation process can be summarized as

\begin{equation}
(X_t, Y_t, M_t, A_t, R_t)
\rightarrow
(X_{t+1}, Y_{t+1}, M_{t+1}, A_{t+1}, R_{t+1}).
\end{equation}

This process shows that up-to-date learning is not only a matter of adding new data or updating parameters. It is a process of keeping the robot's recognition and action relations aligned with the current environment.

\subsection{Integration with the Thinking-Learning Loop}
\label{subsec:adaptation_integration}

The up-to-date adaptation of learning objects is integrated into the thinking-learning loop proposed in Section~\ref{sec:model}. Thinking determines which learning object may need to be updated. Evidence collection provides support for the update. Learning generates candidate changes, and verification prevents unstable or accidental changes from being accepted.

This integration is important because open environments may contain noise, temporary variation, or accidental correlations. Therefore, a new feature, category, model, or action routine should not be accepted only because it appears once. It should be supported by collected evidence and verified through interaction. Once the update is accepted, it becomes part of the robot's knowledge and influences future thinking.

The integration can be described as follows. First, the thinking module evaluates whether the current learning object set $\mathcal{O}_t$ is sufficient for the current observation or task. Second, if an outdated component is detected, the thinking module generates a plan for evidence collection and update verification. Third, the learning module constructs candidate updates from the collected evidence. Fourth, the verification step determines whether the candidate update should be accepted. Finally, the verified update modifies the learning object set and becomes experience for future thinking.

Through repeated interaction, the robot can gradually maintain an up-to-date learning framework. It can discover new features, form new output categories, update learning models, reconstruct action routines, and improve the thinking process that guides these adaptations. In this way, the proposed model supports not only task-level adaptation, but also framework-level adaptation, where the robot learns how its own learning objects should change over time.

\section{Verification}
\label{sec:verification}

\subsection{Experimental Settings}
\label{subsec:experimental_settings}

To verify the effectiveness of the proposed thinking-learning interaction model, four controlled simulation experiments were designed. These experiments were not intended to reproduce a specific physical robot platform. Instead, they were designed to verify whether the proposed model can support up-to-date learning by revising different learning objects, including input features, output categories, learning models, action routines, and thinking strategies.

The first experiment evaluates adaptive input feature discovery. In this experiment, the robot performs a simplified object recognition task involving four object categories: banana, apple, cup, and bottle. At the beginning, the robot is only allowed to use two predefined features, namely shape and size. However, these two features are not always sufficient to distinguish all objects. During interaction, the robot may observe additional candidate features, including color, texture, position, and weight. Among these candidate features, some may be useful for improving recognition, while others may be weak or irrelevant. The purpose of this experiment is to test whether thinking-guided learning can identify useful new features from the candidate feature set when the initial feature set is insufficient.

The second experiment evaluates adaptive output and model expansion. In this experiment, the robot initially recognizes only three known categories: apple, banana, and cup. During testing, samples from a new category, orange, appear in the environment. A conventional closed-set classifier can only force these samples into the existing categories. In contrast, the robot in the proposed model should detect that some samples cannot be explained by the current output set, verify whether they form a stable new category, and then update the output set and learning model. Therefore, this experiment is used to verify whether the proposed model can move beyond predefined output categories and fixed model structures.

The third experiment evaluates adaptive action routine reconstruction. In this experiment, the robot performs a device operation task, where the goal is to set a washing machine to quick-wash mode. The simulated washing machine contains several modes, including standard, heavy, quick, rinse, and spin. The original action routine follows a repeated observation-and-press process: the robot powers on the machine, presses the program button, observes the current mode, judges whether the quick-wash mode is reached, and repeats this process if necessary. Through repeated interaction, the robot may discover that a shorter routine, such as powering on the machine and pressing the program button a fixed number of times before confirmation, can achieve the same goal. This experiment is used to verify whether the proposed model can reconstruct more efficient action routines from interaction history.

The fourth experiment evaluates learning-enhanced thinking. In this experiment, the robot repeatedly selects evidence for different learning tasks, including object recognition, action routine reconstruction, and new category discovery. The available evidence sources include color, shape, texture, position, historical successful action sequences, historical failed action sequences, unknown sample clusters, random noise, past similar tasks, and partial input matches. At the beginning, the robot may not know which evidence sources are useful for different tasks. After repeated learning and verification, it should improve its ability to select useful evidence and avoid repeated invalid choices. This experiment directly verifies whether learning results can improve future thinking.

The four experiments and their compared methods are summarized in Table~\ref{tab:verification_scenarios}. Each experiment was repeated for 30 independent rounds. The result of each round was written into a data file after completion. A completion flag was also recorded, so that the program could resume from the last incomplete round if it was interrupted.

\begin{table*}[htp]
\centering
\caption{Verification scenarios and compared methods.}
\label{tab:verification_scenarios}
\begin{tabular}{p{0.20\linewidth} p{0.34\linewidth} p{0.34\linewidth}}
\toprule
Scenario & Experimental Purpose & Compared Methods \\
\midrule
Input feature adaptation
& Test whether the robot can discover useful new features when predefined features are insufficient.
& Fixed-feature learning; random feature expansion; proposed thinking-guided feature adaptation. \\

Output and model adaptation
& Test whether the robot can detect unknown samples, form a new category, and update the learning model.
& Closed-set classifier; open-set detection only; random category expansion; proposed thinking-learning model. \\

Action routine reconstruction
& Test whether the robot can reconstruct a shorter and more reliable action routine from interaction history.
& Fixed routine; random action search; trial-and-error RL-like baseline; proposed thinking-learning model. \\

Learning-enhanced thinking
& Test whether previous learning results can improve future evidence selection and reasoning.
& No thinking improvement; memory-only retrieval; proposed learning-enhanced thinking. \\
\bottomrule
\end{tabular}
\end{table*}

\subsection{Compared Methods}
\label{subsec:compared_methods}

Different baselines were selected according to the purpose of each experiment.

For adaptive input feature discovery, three methods were compared. The fixed-feature method only uses the initial predefined features and does not allow new features to be added. The random feature expansion method randomly selects candidate features and adds them into the feature set. The proposed method uses the thinking module to analyze recognition errors, select candidate features, collect evidence, and verify whether the selected feature is useful.

For adaptive output and model expansion, four methods were compared. The closed-set classifier always assigns samples to existing categories and cannot detect unknown samples. The open-set detection-only method can detect samples that do not fit existing categories, but it does not form a new category or update the model. The random category expansion method randomly decides whether to create a new category. The proposed method uses thinking-guided verification to judge whether unknown samples form a stable new category and then updates the output set and learning model.

For adaptive action routine reconstruction, four methods were compared. The fixed routine always follows the original predefined action sequence. The random action search method randomly explores possible action sequences. The trial-and-error RL-like baseline searches for a successful sequence through repeated attempts. The proposed method analyzes successful historical action sequences, infers a shorter candidate routine, and verifies whether it can reliably achieve the same task goal.

For learning-enhanced thinking, three methods were compared. The no-improvement method selects evidence without updating its selection strategy. The memory-only method can retrieve historical records, but it does not learn which evidence sources are more useful. The proposed method updates its evidence selection strategy according to previous learning and verification results.

\subsection{Results of Adaptive Input Feature Discovery}
\label{subsec:feature_results}

The first experiment evaluates whether the robot can discover useful new input features when the initial feature set is insufficient. In this experiment, the main metrics include accuracy after adaptation, useful feature discovery rate, false feature acceptance, evidence cost, and adaptation steps. Accuracy measures the final recognition performance. Useful feature discovery rate measures whether the method successfully identifies the truly useful feature. False feature acceptance measures how many ineffective features are wrongly accepted. Evidence cost measures the amount of evidence collected for feature verification. Adaptation steps measure how many update attempts are needed before the method reaches a stable feature set.

The results are shown in Table~\ref{tab:feature_results}. The fixed-feature method achieved only 0.419 accuracy because it could not use any new feature. Random feature expansion improved the accuracy to 0.704, showing that feature expansion is useful in the open environment. However, the proposed method further improved the accuracy to 0.845. Compared with random feature expansion, the proposed method improved the final accuracy by about 14.1 percentage points.

\begin{table}[htp]
\centering
\caption{Results of adaptive input feature discovery.}
\label{tab:feature_results}
\begin{tabular}{lcccc}
\toprule
Method & Acc. & Disc. & False & Steps \\
\midrule
Fixed feature & 0.419 & 0.000 & 0.000 & 0.0 \\
Random expansion & 0.704 & 0.633 & 1.833 & 2.4 \\
Proposed & 0.845 & 0.933 & 0.300 & 1.5 \\
\bottomrule
\end{tabular}
\end{table}

Here, ``Acc.'' denotes the accuracy after adaptation, ``Disc.'' denotes the useful feature discovery rate, ``False'' denotes the number of false feature acceptances, and ``Steps'' denotes the average number of adaptation steps. The proposed method achieved a useful feature discovery rate of 0.933, while random expansion achieved only 0.633. In addition, the false feature acceptance was reduced from 1.833 to 0.300. These results show that thinking-guided evidence selection can identify useful features more reliably than random feature expansion.

\begin{figure*}[htp]
\centering
\begin{minipage}{0.45\linewidth}
\centering
\includegraphics[width=\linewidth]{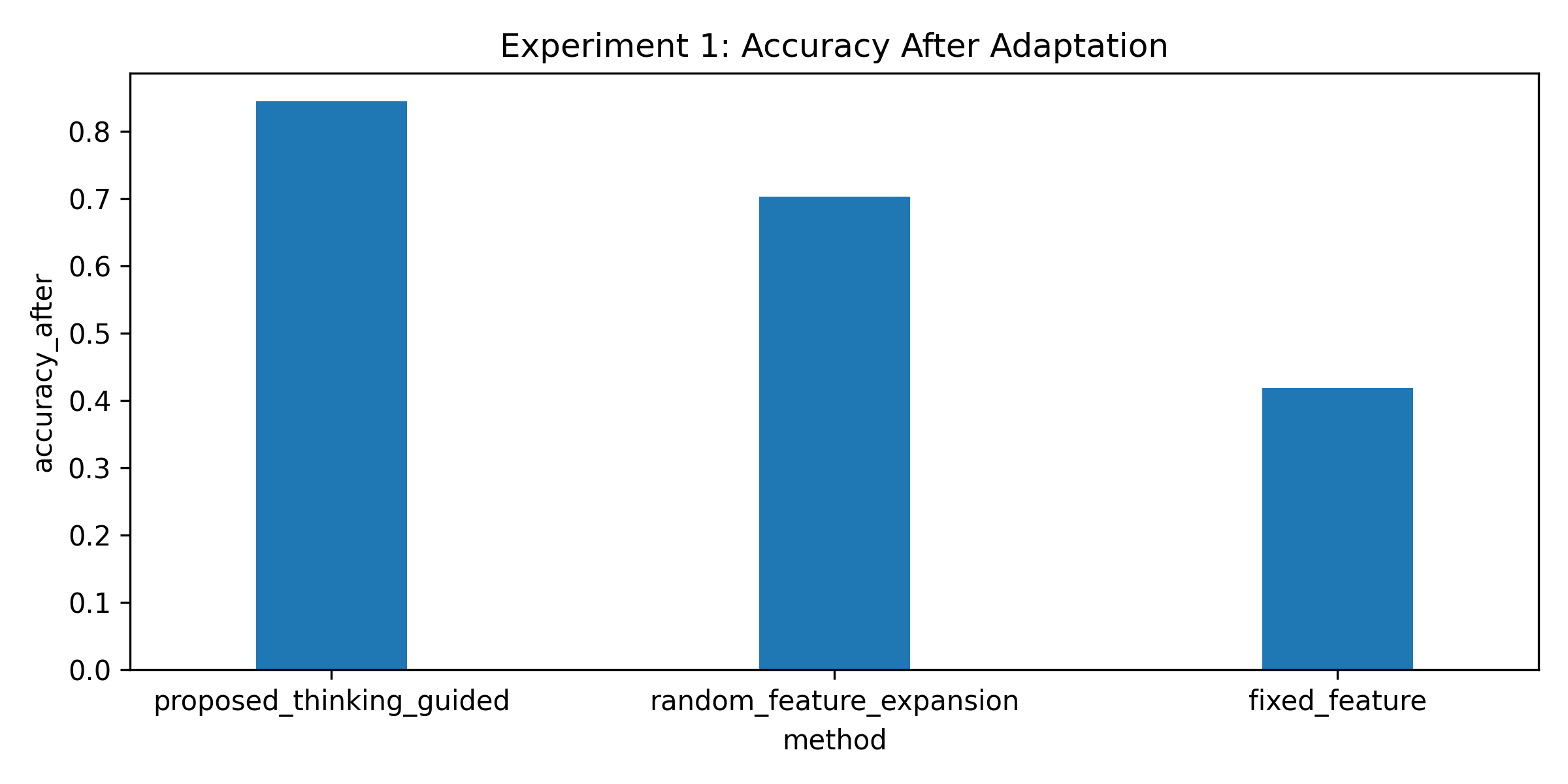}

(a) Accuracy after adaptation.
\end{minipage}
\hfill
\begin{minipage}{0.45\linewidth}
\centering
\includegraphics[width=\linewidth]{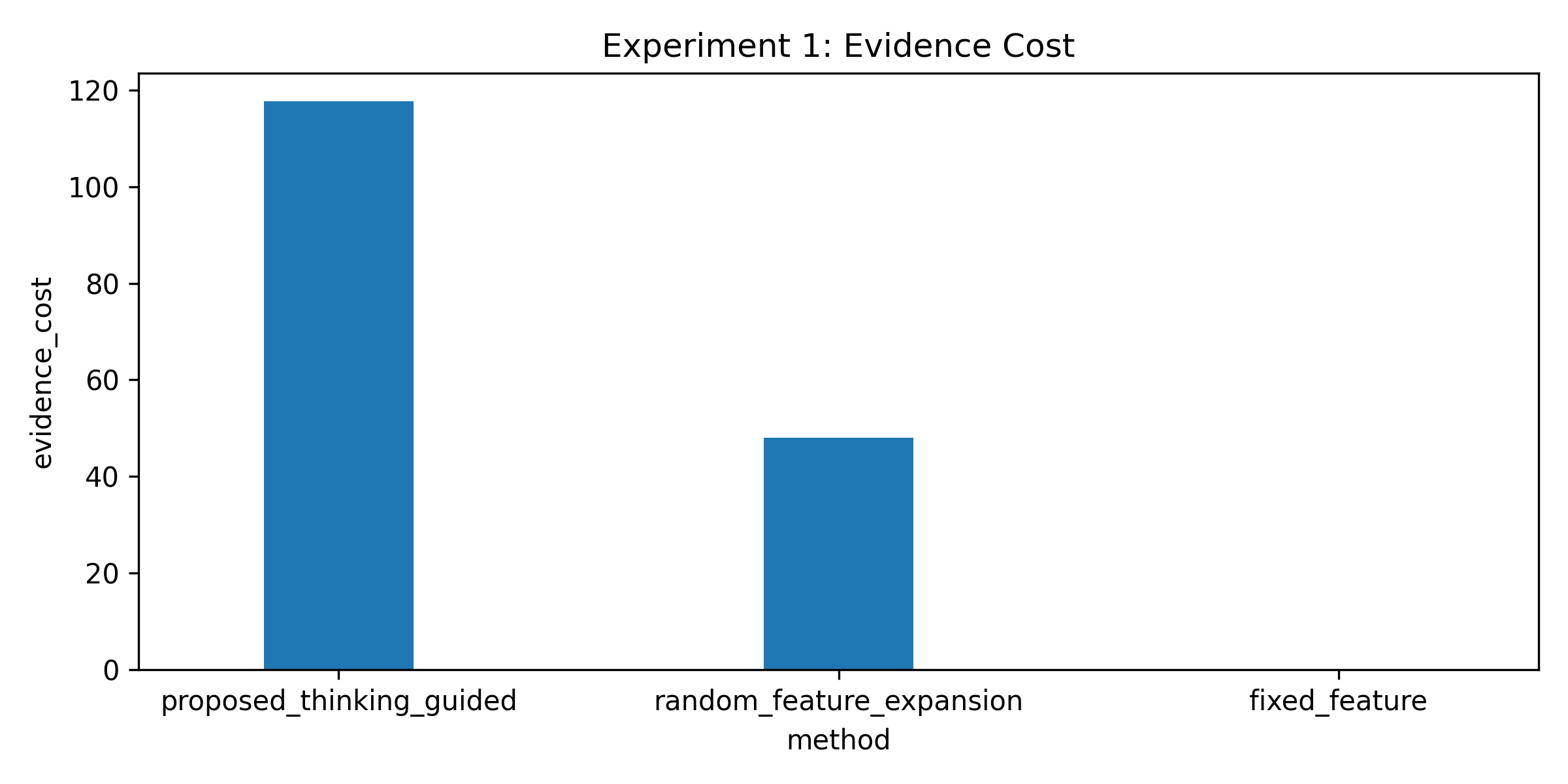}

(b) Evidence cost.
\end{minipage}
\caption{Results of adaptive input feature discovery. The proposed method achieves higher final accuracy by selecting and verifying more useful features.}
\label{fig:feature_results}
\end{figure*}

Fig.~\ref{fig:feature_results} further compares the accuracy and evidence cost. The proposed method achieved the highest accuracy, but its evidence cost was also higher than that of random feature expansion. This is because the proposed method collected and verified more evidence before accepting a new feature. Therefore, the advantage of the proposed method is not simply lower evidence cost, but higher adaptation reliability and fewer false feature acceptances.

\subsection{Results of Adaptive Output and Model Expansion}
\label{subsec:output_model_results}

The second experiment evaluates whether the robot can detect new categories and update the learning model. The main metrics include unknown detection rate, new category formation accuracy, false category creation rate, model update success rate, and forgetting rate. Unknown detection rate measures whether unknown samples can be separated from known categories. New category formation accuracy measures whether the method correctly adds the true new category. False category creation rate measures whether the method incorrectly creates unnecessary categories. Model update success rate measures whether the updated model can recognize both old and new categories. Forgetting rate measures the performance loss on old categories after model update.

Table~\ref{tab:output_model_results} shows the results. The closed-set classifier had no ability to detect unknown samples, and its unknown detection rate was 0.000. The open-set detection-only method achieved a high unknown detection rate of 0.997, but it could not form a new category or update the model. Random category expansion also had the same unknown detection rate, but its new category formation accuracy was only 0.167, and its false category creation rate was 0.367.

\begin{table}[htp]
\centering
\caption{Results of adaptive output and model expansion.}
\label{tab:output_model_results}
\begin{tabular}{lccccc}
\toprule
Method & Unk. & New & False & Model & Forget \\
\midrule
Closed-set & 0.000 & 0.000 & 0.000 & 0.000 & 0.000 \\
Open-set only & 0.997 & 0.000 & 0.000 & 0.000 & 0.000 \\
Random expansion & 0.997 & 0.167 & 0.367 & 0.167 & 0.001 \\
Proposed & 0.997 & 1.000 & 0.000 & 1.000 & 0.003 \\
\bottomrule
\end{tabular}
\end{table}

Here, ``Unk.'' denotes the unknown detection rate, ``New'' denotes the new category formation accuracy, ``False'' denotes the false category creation rate, ``Model'' denotes the model update success rate, and ``Forget'' denotes the forgetting rate. The proposed method achieved 1.000 new category formation accuracy and 1.000 model update success rate. Its false category creation rate was 0.000, which means that the thinking-guided verification process prevented incorrect category creation. The forgetting rate was only 0.003, indicating that the updated model maintained the performance of old categories while learning the new category.

\begin{figure*}[htp]
\centering
\begin{minipage}{0.45\linewidth}
\centering
\includegraphics[width=\linewidth]{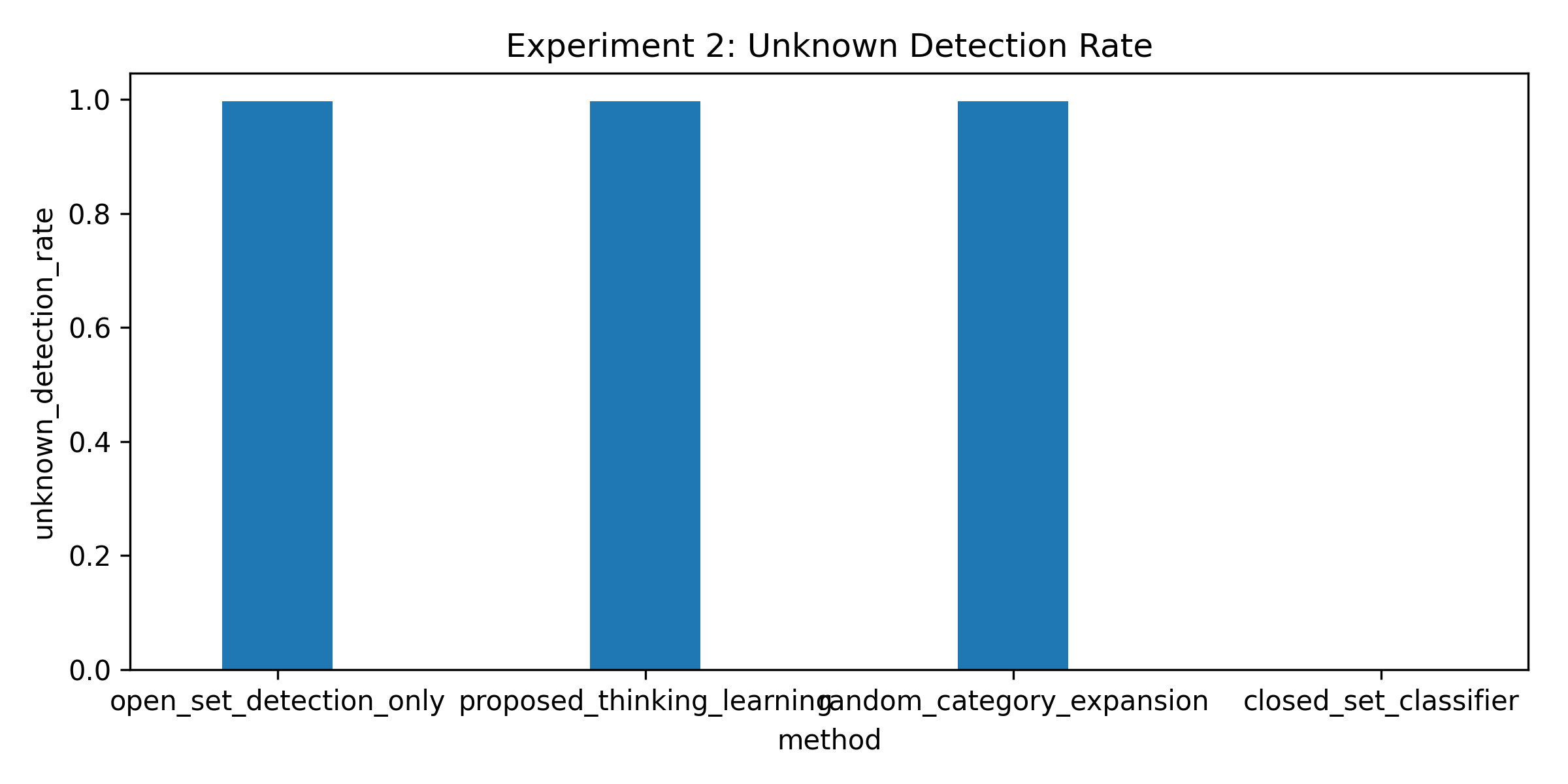}

(a) Unknown detection rate.
\end{minipage}
\hfill
\begin{minipage}{0.45\linewidth}
\centering
\includegraphics[width=\linewidth]{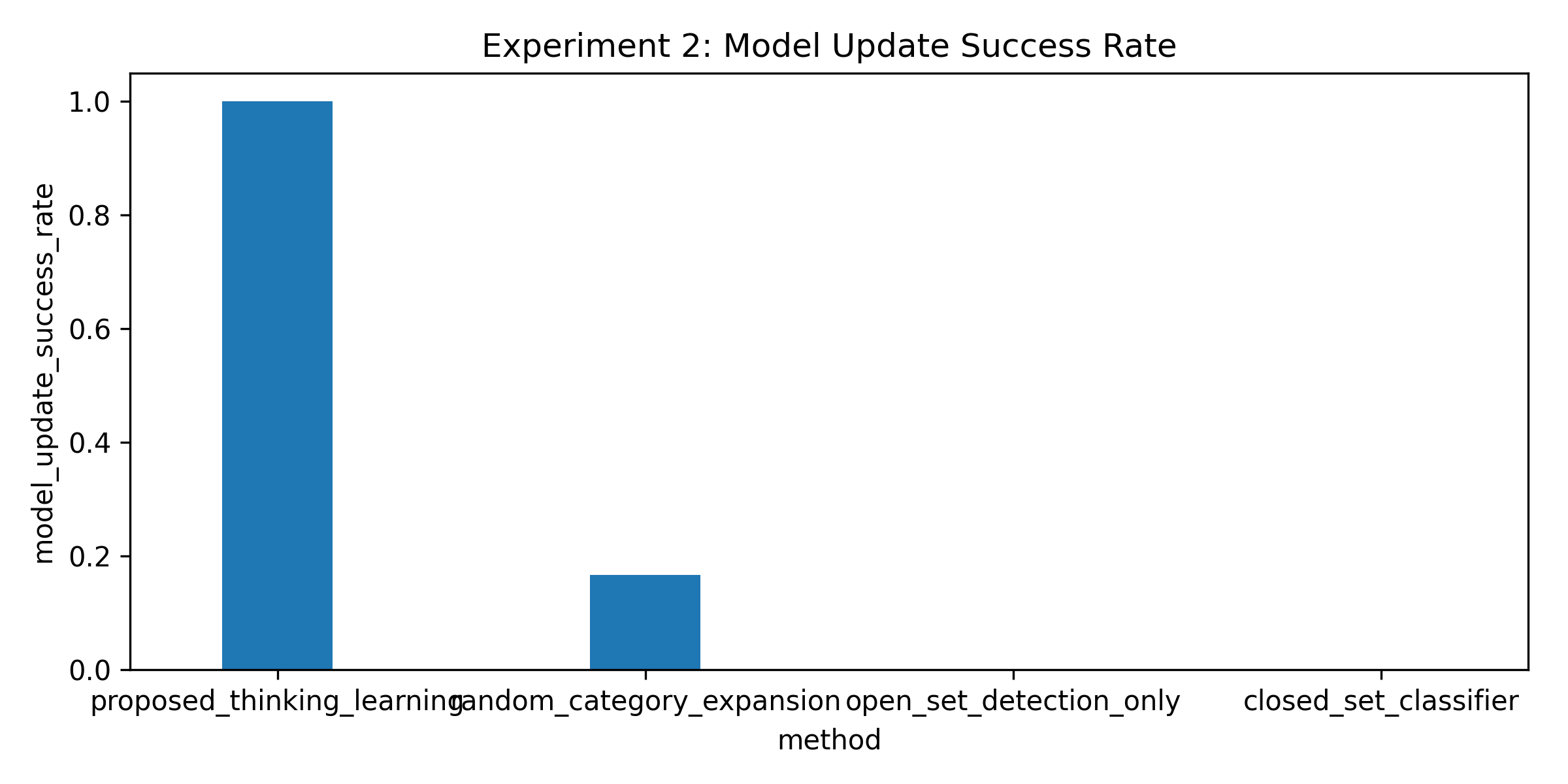}

(b) Model update success rate.
\end{minipage}
\caption{Results of adaptive output and model expansion. The proposed method can convert unknown samples into a verified new category and update the model successfully.}
\label{fig:output_model_results}
\end{figure*}

Fig.~\ref{fig:output_model_results} shows the unknown detection rate and model update success rate. Although the open-set detection-only method could detect unknown samples, it could not transform the detected samples into a new output category. In contrast, the proposed method not only detected unknown samples, but also verified them, formed a new output category, and updated the model. This supports the claim that the proposed method can move beyond predefined outputs and models.

\subsection{Results of Adaptive Action Routine Reconstruction}
\label{subsec:action_results}

The third experiment evaluates whether the robot can reconstruct a more efficient action routine. The task was to set a washing machine to quick-wash mode. The original routine followed a repeated observation-and-press process, while the proposed method attempted to infer a shorter routine from successful historical executions.

The main metrics include task success rate, average action length, adaptation time, routine compression ratio, and failed trial rate. Task success rate measures whether the routine can complete the target task. Average action length measures the efficiency of the learned routine. Adaptation time measures how many trials or historical executions are needed before a new routine is obtained. Routine compression ratio measures how much the original routine is shortened. Failed trial rate measures the proportion of unsuccessful attempts during adaptation.

The results are shown in Table~\ref{tab:action_results}. The fixed routine failed to complete the task in this simulation, with a task success rate of 0.000. Random action search achieved only 0.067 success rate and had a high failed trial rate of 0.933. The trial-and-error RL-like method achieved 1.000 success rate and reduced the average action length to 4.0. The proposed method also achieved 1.000 success rate and an average action length of 4.0.

\begin{table}[htp]
\centering
\caption{Results of adaptive action routine reconstruction.}
\label{tab:action_results}
\begin{tabular}{lccccc}
\toprule
Method & Succ. & Len. & Time & Comp. & Fail \\
\midrule
Fixed routine & 0.000 & 13.0 & 0.0 & 0.000 & 0.000 \\
Random search & 0.067 & 4.27 & 10.0 & 0.671 & 0.933 \\
RL-like & 1.000 & 4.0 & 3.9 & 0.692 & 0.293 \\
Proposed & 1.000 & 4.0 & 5.0 & 0.692 & 0.000 \\
\bottomrule
\end{tabular}
\end{table}

Here, ``Succ.'' denotes the task success rate, ``Len.'' denotes the average action length, ``Time'' denotes the adaptation time, ``Comp.'' denotes the routine compression ratio, and ``Fail'' denotes the failed trial rate. The proposed method reduced the average action length from 13.0 to 4.0, with a compression ratio of 0.692. Compared with random action search, the proposed method improved the task success rate from 0.067 to 1.000 and reduced the failed trial rate from 0.933 to 0.000.

\begin{figure*}[htp]
\centering
\begin{minipage}{0.45\linewidth}
\centering
\includegraphics[width=\linewidth]{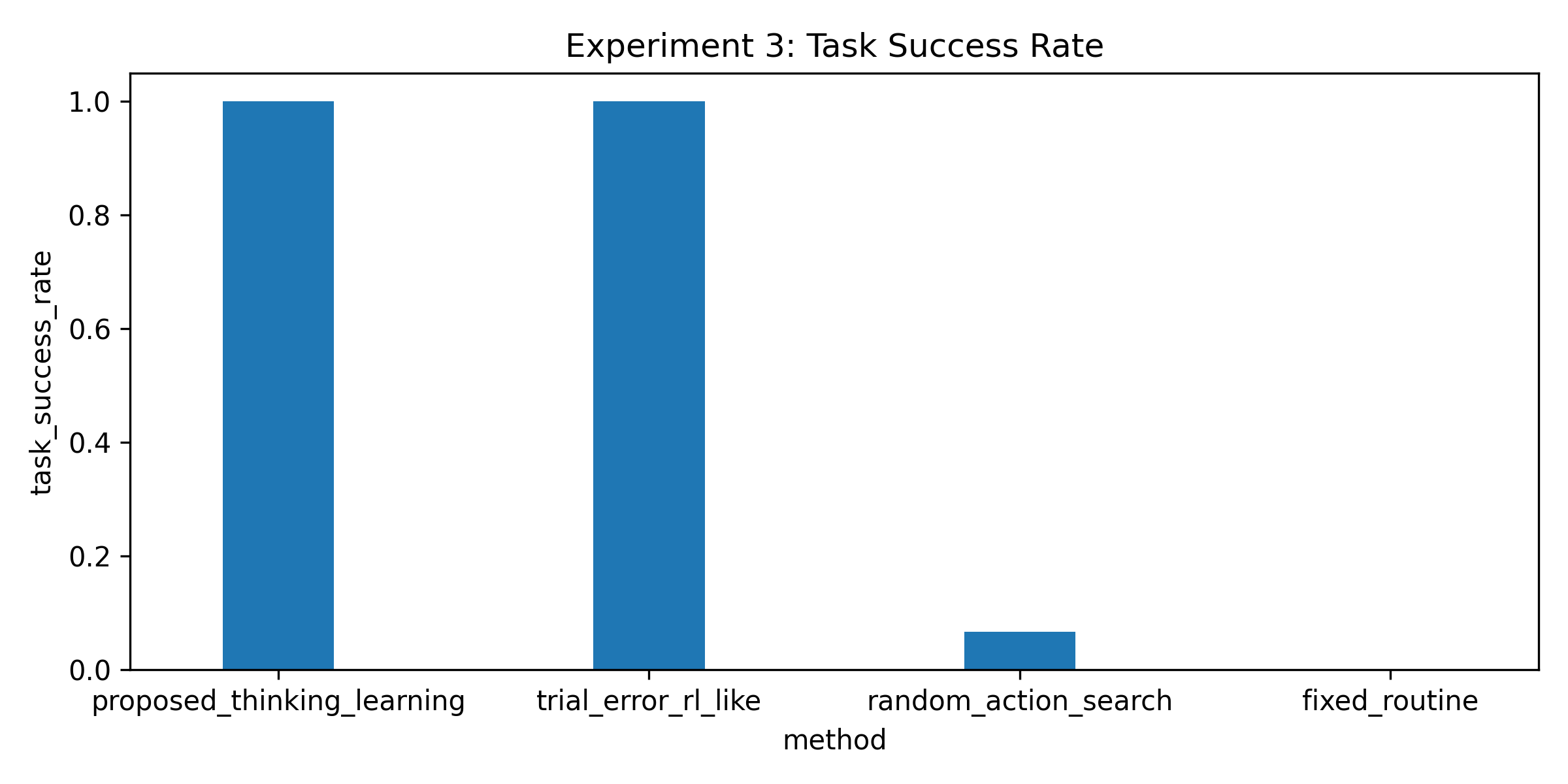}

(a) Task success rate.
\end{minipage}
\hfill
\begin{minipage}{0.45\linewidth}
\centering
\includegraphics[width=\linewidth]{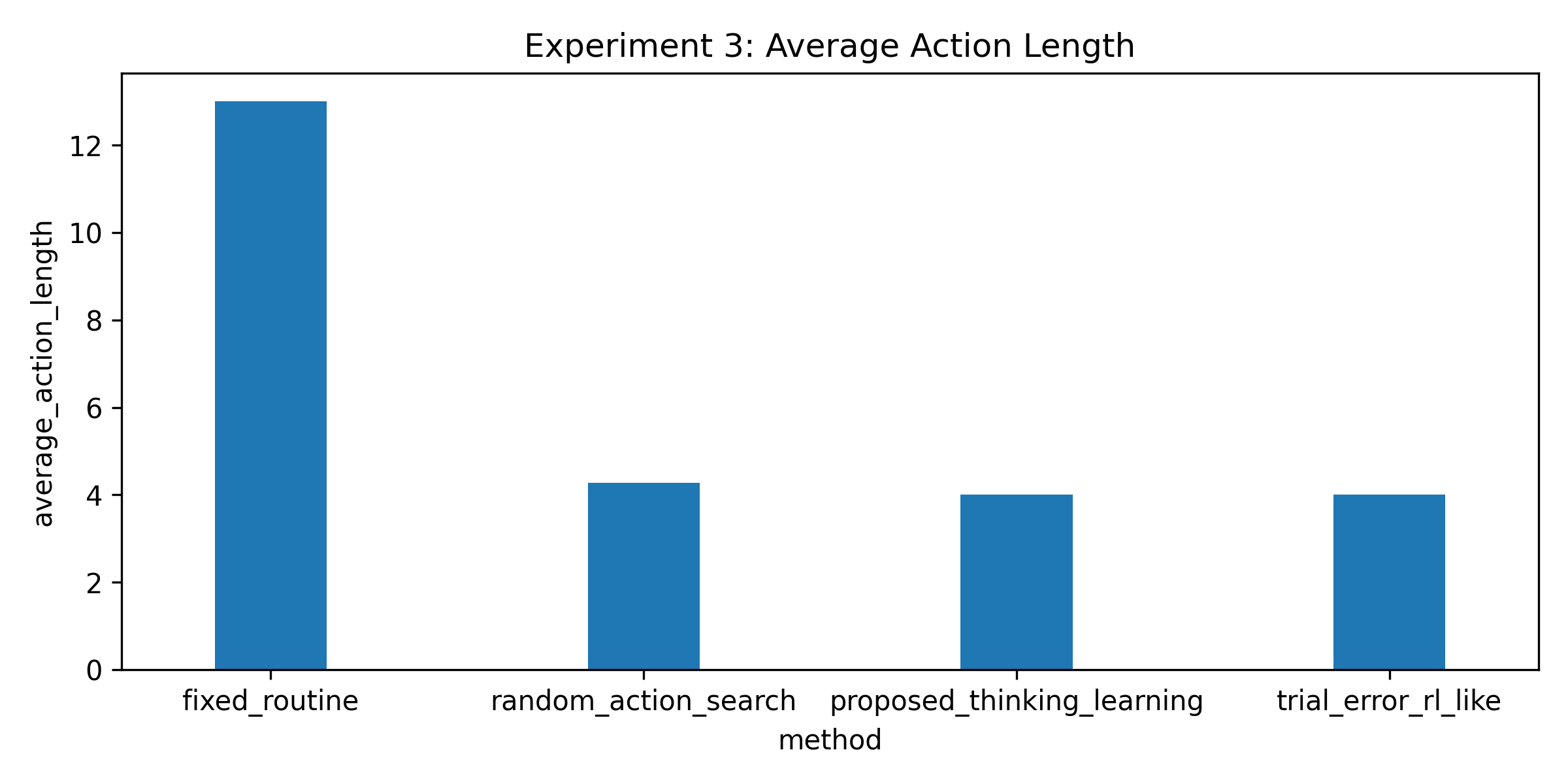}

(b) Average action length.
\end{minipage}
\caption{Results of adaptive action routine reconstruction. The proposed method achieves stable task completion and compresses the original action routine.}
\label{fig:action_results}
\end{figure*}

Fig.~\ref{fig:action_results} shows the task success rate and average action length. The proposed method achieved the same final action length as the trial-and-error RL-like baseline. However, the proposed method had no failed trials, while the RL-like method had a failed trial rate of 0.293. Therefore, the advantage of the proposed method in this scenario is not only the final routine length, but also the evidence-guided and interpretable reconstruction process. The robot reconstructs the routine by analyzing successful historical action sequences rather than by relying only on trial-and-error exploration.

\begin{figure}[htp]
\centering
\includegraphics[width=0.95\linewidth]{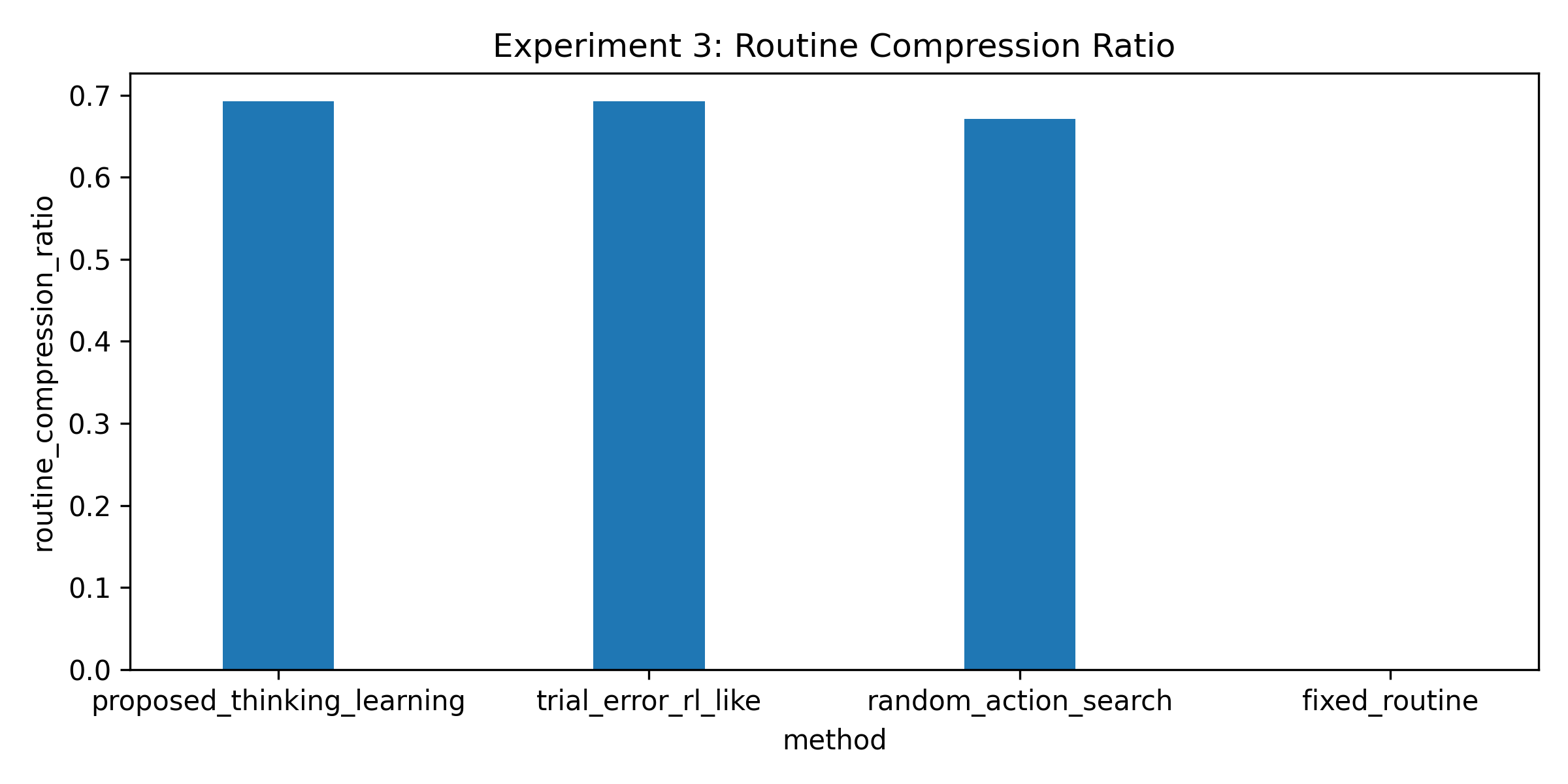}
\caption{Routine compression ratio in the action routine reconstruction experiment.}
\label{fig:action_compression}
\end{figure}

Fig.~\ref{fig:action_compression} further shows the routine compression ratio. The proposed method achieved a compression ratio of 0.692, which means that the original action routine was substantially shortened. This supports the claim that the proposed model can help robots move beyond predefined action routines.

\subsection{Results of Learning-Enhanced Thinking}
\label{subsec:thinking_results}

\begin{figure*}[hpt]
\centering
\begin{minipage}{0.45\linewidth}
\centering
\includegraphics[width=\linewidth]{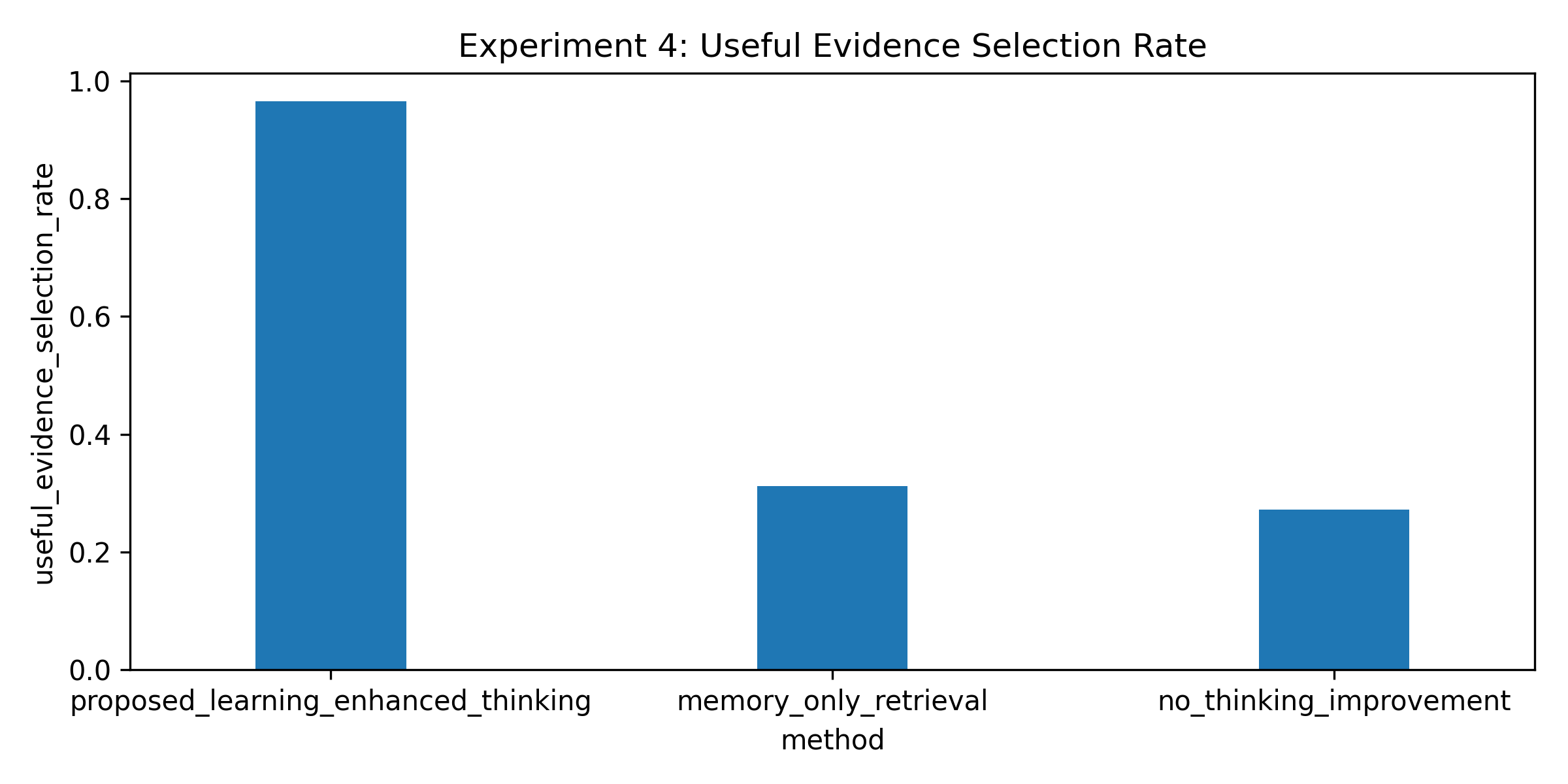}

(a) Useful evidence selection rate.
\end{minipage}
\hfill
\begin{minipage}{0.45\linewidth}
\centering
\includegraphics[width=\linewidth]{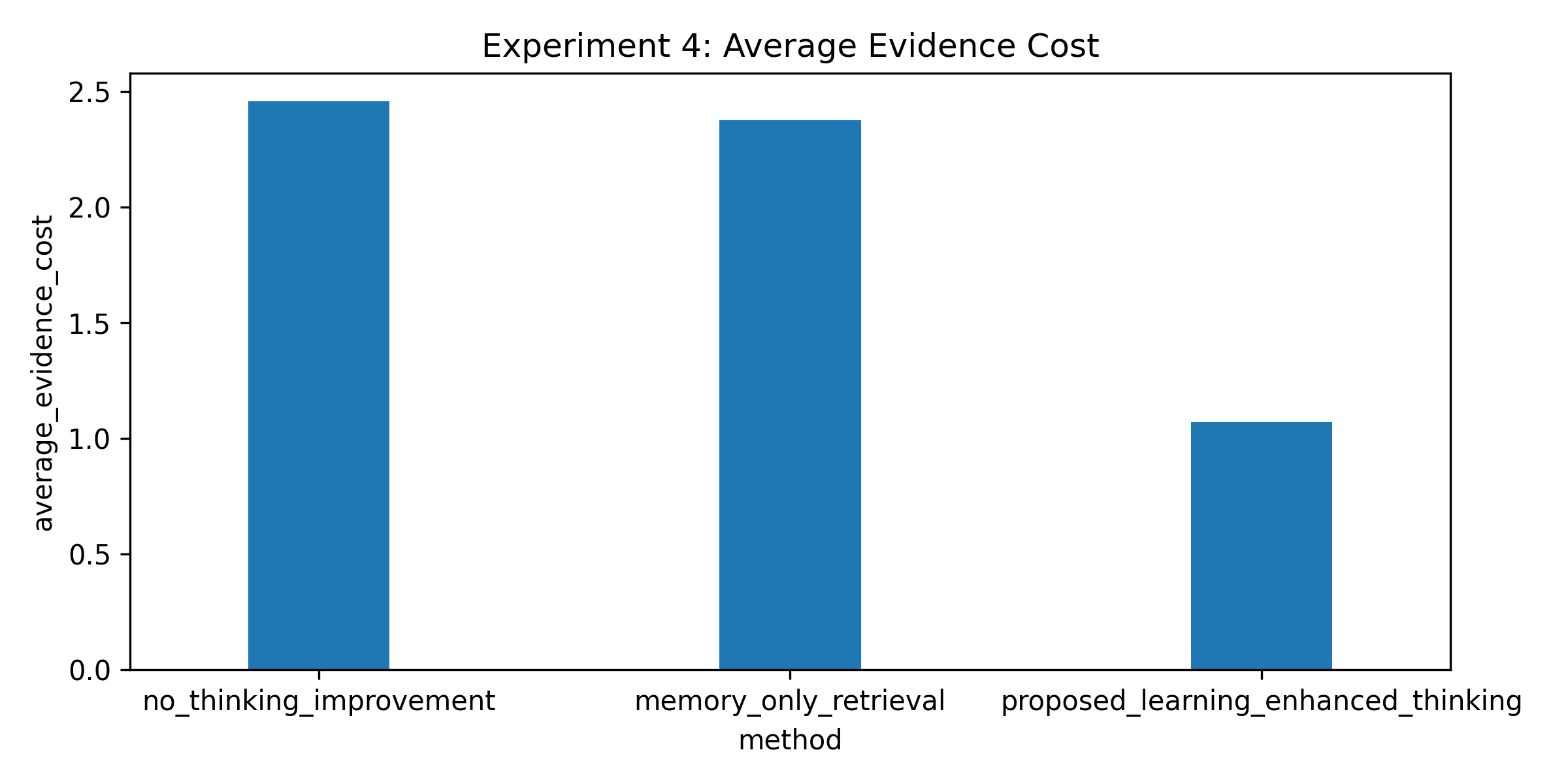}

(b) Average evidence cost.
\end{minipage}
\caption{Results of learning-enhanced thinking. The proposed method improves evidence selection and reduces evidence cost.}
\label{fig:thinking_results}
\end{figure*}

The fourth experiment evaluates whether learning results can improve future thinking. The task was to select useful evidence for different learning problems. Without thinking improvement, the robot selected evidence almost randomly. With memory-only retrieval, the robot could use historical information, but it did not update its evidence selection strategy. The proposed method used previous learning results to improve future evidence selection.

The main metrics include useful evidence selection rate, average evidence cost, repeated error rate, thinking success rate, and scope expansion success rate. Useful evidence selection rate measures whether the thinking module selects evidence that is actually helpful for the learning task. Average evidence cost measures the cost of obtaining useful evidence. Repeated error rate measures whether the robot repeatedly selects invalid evidence. Thinking success rate measures whether the thinking module generates a useful evidence selection plan. Scope expansion success rate measures whether searching broader historical or contextual evidence helps the current learning task.

Table~\ref{tab:thinking_results} shows the results. Without thinking improvement, the useful evidence selection rate was only 0.272. Memory-only retrieval slightly improved it to 0.312. In contrast, the proposed method achieved a useful evidence selection rate of 0.965. This shows that learning results can significantly improve the robot's ability to select useful evidence.

\begin{table}[htp]
\centering
\caption{Results of learning-enhanced thinking.}
\label{tab:thinking_results}
\begin{tabular}{lccccc}
\toprule
Method & Useful & Cost & Repeat & Think & Scope \\
\midrule
No improvement & 0.272 & 2.456 & 0.408 & 0.272 & 0.129 \\
Memory-only & 0.312 & 2.376 & 0.380 & 0.312 & 0.129 \\
Proposed & 0.965 & 1.069 & 0.011 & 0.965 & 0.965 \\
\bottomrule
\end{tabular}
\end{table}

Here, ``Useful'' denotes the useful evidence selection rate, ``Cost'' denotes the average evidence cost, ``Repeat'' denotes the repeated error rate, ``Think'' denotes the thinking success rate, and ``Scope'' denotes the scope expansion success rate. The proposed method reduced the average evidence cost from 2.456 to 1.069. It also reduced the repeated error rate from 0.408 to 0.011. Moreover, the scope expansion success rate increased from 0.129 to 0.965, indicating that the robot learned how to search broader historical or contextual evidence more effectively.

Fig.~\ref{fig:thinking_results} shows the useful evidence selection rate and evidence cost. The proposed method selected useful evidence much more frequently than the baselines and required less average evidence cost. This result supports the core claim of learning-enhanced thinking: the robot does not only learn task-specific knowledge, but also learns how to think more effectively about future learning problems.

\subsection{Discussion}
\label{subsec:verification_discussion}

The experimental results show that the proposed thinking-learning interaction model provides advantages in adaptive robot learning. In input feature adaptation, the proposed method improved the final recognition accuracy from 0.419 under fixed features to 0.845, and achieved a useful feature discovery rate of 0.933. In output and model adaptation, the proposed method achieved 1.000 new category formation accuracy and 1.000 model update success rate, while random category expansion achieved only 0.167 on both metrics. In action routine reconstruction, the proposed method reduced the average action length from 13.0 to 4.0 and achieved a routine compression ratio of 0.692. In learning-enhanced thinking, the useful evidence selection rate increased from 0.272 to 0.965, and the repeated error rate decreased from 0.408 to 0.011.

These results indicate that the proposed model can help robots move beyond predefined learning objects. The thinking module helps the robot identify what should be learned and which evidence should be collected. The learning module then transforms the collected evidence into candidate updates. Verified learning results are used not only to improve task performance, but also to improve future thinking strategies.

It should also be noted that the advantages of the proposed method appear in different forms across different scenarios. In feature adaptation, the proposed method achieves higher accuracy and fewer false feature acceptances, but it requires more evidence collection than random expansion. In action routine reconstruction, the proposed method achieves the same final action length as the trial-and-error RL-like method, but it avoids failed trials and provides a more interpretable evidence-guided update process. Therefore, the proposed method should be understood as a framework for reliable and interpretable adaptation, rather than merely a method for minimizing short-term exploration cost.

\section{Conclusion}
\label{sec:conclusion}

This paper proposed a thinking-learning interaction model for up-to-date autonomous robot learning in open and changing environments. Different from conventional learning pipelines that mainly optimize model parameters under predefined inputs, outputs, models, and action routines, the proposed model treats these learning objects themselves as adaptive components. The core idea is that thinking guides learning by identifying learning needs, selecting evidence, organizing learning materials, and planning verification actions, while learning enhances thinking by updating task knowledge, evidence-selection experience, action strategies, and future reasoning processes.

Based on this bidirectional mechanism, the robot can gradually move beyond predefined learning settings. The proposed model supports adaptive input feature discovery, output category expansion, learning model update, and action routine reconstruction. It also enables the robot to learn how to select useful evidence and how to expand the evidence collection scope during long-term interaction.

The verification results show that the proposed model provides clear advantages in adaptive learning. In feature adaptation, the proposed method improves the final recognition accuracy from 0.419 to 0.845. In output and model adaptation, it achieves 1.000 new-category formation accuracy and 1.000 model-update success rate. In action routine reconstruction, it reduces the average action length from 13.0 to 4.0. In learning-enhanced thinking, the useful evidence selection rate increases from 0.272 to 0.965. These results indicate that the proposed model can improve both task-level adaptation and reasoning-level adaptation.

Future work will extend the proposed model to more complex real-world robotic tasks, including visual perception, multi-step manipulation, and long-term human-robot interaction. In addition, more detailed mechanisms for updating the thinking module, such as memory retrieval, reinforcement learning, and fine-tuning of large language models, will be further investigated.


\ifCLASSOPTIONcaptionsoff
  \newpage
\fi

\bibliographystyle{IEEEtran}
\bibliography{ref}

%

\begin{IEEEbiography}{Hong Su}
  received the MS and PhD degrees, in 2006 and 2022, respectively, from Sichuan University, Chengdu, China. He is currently a researcher of Chengdu University of Information Technology Chengdu, China. His research interests include blockchain, cross-chain and smart contract.
\end{IEEEbiography}




\end{document}